\documentclass[journal]{IEEEtran}
\usepackage[small]{caption}
\usepackage{epsfig}
\usepackage{stfloats}
\usepackage{bbm}
\usepackage{subfig}

\usepackage{times}
\usepackage{soul}
\usepackage{url}
\usepackage[hidelinks]{hyperref}

\usepackage{graphicx}
\usepackage{amsmath}
\usepackage{amsthm}
\usepackage{booktabs}
\usepackage{algorithmic}

\usepackage{epstopdf}
\usepackage{bm}
\usepackage{color}
\usepackage{multirow}
\usepackage[ruled]{algorithm2e}

\begin{document}

\title{Expert Training: Task Hardness Aware Meta-Learning for Few-Shot Image Classification}

\author{Yucan~Zhou,
        Yu~Wang,
        Jianfei~Cai,
        Yu~Zhou,
        Qinghua~Hu,~\IEEEmembership{Senior~Member,~IEEE,}
        and Weiping~Wang
\thanks{Y.~Zhou, Y.~Zhou, and W.~Wang are with the Institute of Information Engineering, Chinese Academy of Sciences, Beijing 100195, China (e-mail: zhouyucan@iie.ac.cn, zhouyu@iie.ac.cn, wangweiping@iie.ac.cn).}
\thanks{Y.~Wang, Q.~Hu are with the College of Intelligence and Computing, Tianjin University, Tianjin 300350, China (e-mail: armstrong\_wangyu@tju.edu.cn, huqinghua@tju.edu.cn).}
\thanks{J. Cai is with the Department of Data Science \& AI, Monash University, Victoria 3800, Australia (e-mail: jianfei.cai@monash.edu).}
}

%

\maketitle

\begin{abstract}
  Deep neural networks are highly effective when a large number of labeled samples are available but fail with few-shot classification tasks. Recently, meta-learning methods have received much attention, which train a meta-learner on massive additional tasks to gain the knowledge to instruct the few-shot classification. Usually, the training tasks are randomly sampled and performed indiscriminately, often making the meta-learner stuck into a bad local optimum. Some works in the optimization of deep neural networks have shown that a better arrangement of training data can make the classifier converge faster and perform better. Inspired by this idea, we propose an easy-to-hard expert meta-training strategy to arrange the training tasks properly, where easy tasks are preferred in the first phase, then, hard tasks are emphasized in the second phase. A task hardness aware module is designed and integrated into the training procedure to estimate the hardness of a task based on the distinguishability of its categories. In addition, we explore multiple hardness measurements including the semantic relation, the pairwise Euclidean distance, the Hausdorff distance, and the Hilbert-Schmidt independence criterion. Experimental results on the miniImageNet and tieredImageNetSketch datasets show that the meta-learners can obtain better results with our expert training strategy.
\end{abstract}


\ifCLASSOPTIONpeerreview
\begin{center} \bfseries EDICS Category: 3-BBND \end{center}
\fi
\IEEEpeerreviewmaketitle

\section{Introduction}
Recently, deep learning methods have achieved great success with the benefit of large-scale labeled data \cite{deng2009imagenet,taigman2014deepface,miech2019howto100m,lin2014microsoft}, such as ImageNet, which consists of $1,000$ categories with thousands of images in each category. However, in the real world scenarios, samples in many categories are rarely seen or expensive to collect (e.g., in the medical diagnosis and the hyperspectral image processing areas \cite{mondal2018few,tanner2018generative,yuan2019hyperspectral}), which leads to the challenge of training big models with very few data. Therefore, few-shot learning (FSL) has been introduced to recognize categories with only several training samples \cite{hariharan2016low,li2019large}.

\begin{figure*}[!bp]
\setlength{\abovecaptionskip}{0pt}
\centering
\includegraphics[width=0.85\textwidth]{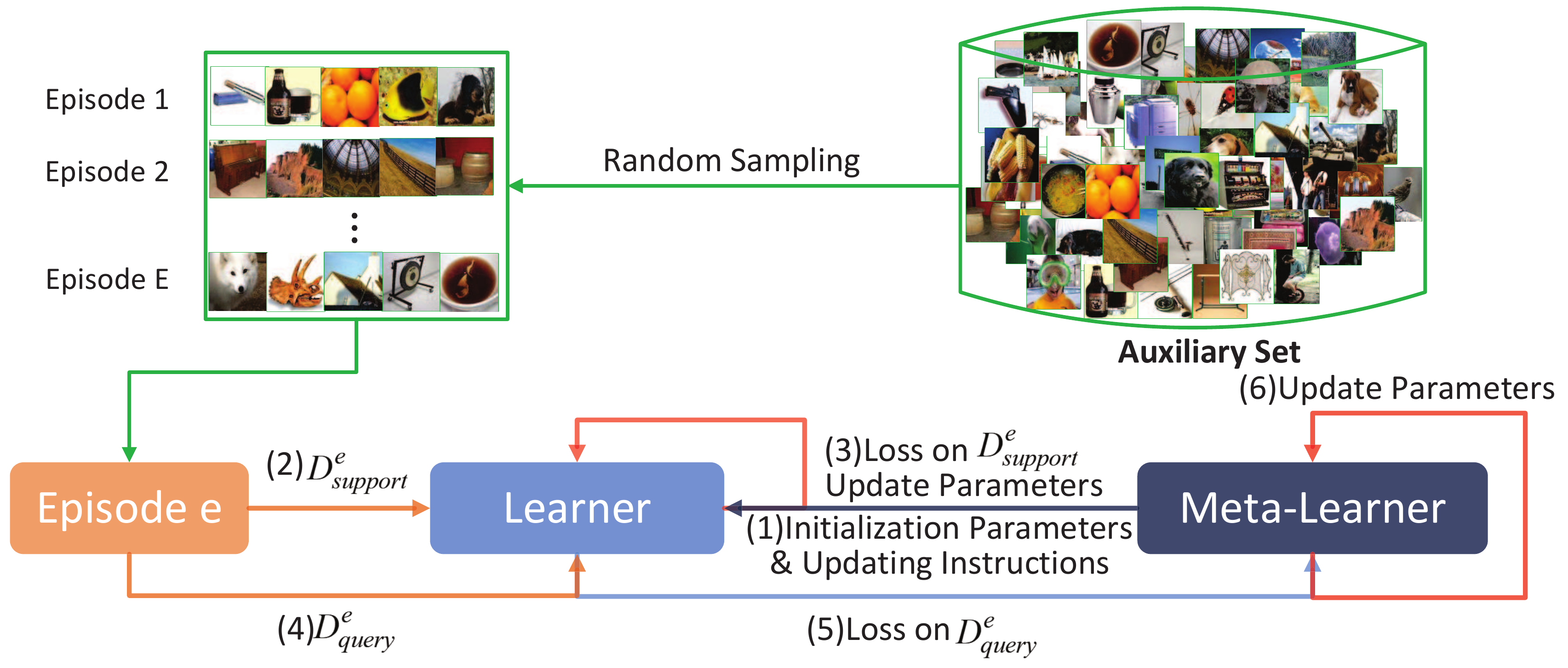}
\caption{An illustration of the episode-based training. During training, for an episode $e=\{D^e_{support}, D^e_{query}\}$, the following operations are performed sequentially: (1) the learner receives parameters and updating instructions from the meta-learner to initialize its parameters; (2) the support samples $D^e_{support}$ are provided to the learner; (3) the learner updates its parameters to minimize the loss on $D^e_{support}$ under the updating instructions; (4) the query samples $D^e_{query}$ are fed to the updated learner to compute the query loss; (5) the loss on $D^e_{query}$ is propagated to the meta-learner; (6) the meta-learner updates its parameters to minimize the obtained loss. Finally, the meta-learner produces new initialization parameters and updating instructions for the next episode.}
\label{fig:episode}
\end{figure*}

Meta-learning is one of the most successful FSL techniques. Different from the traditional recognition models, where a classifier is trained on the training data and aims to generalize well on unseen samples from a similar data distribution, meta-learning methods are trained with many homogeneous tasks to learn the meta-knowledge, which can inversely instruct the training procedure of classifiers for unseen tasks from a similar task distribution \cite{sung2018learningToLearn,ravi2017optimization}. They usually follow an episode-based training strategy, where the episodes are randomly sampled from an auxiliary dataset with categories excluded to those in the novel unseen tasks. To mimic what happens during inference, $N$ categories are included in each episode, and $K$ support samples and $Q$ query samples are selected for each category \cite{ravi2017optimization,sung2018learning,li2017meta,finn2017model}. Each episode has a specific learner and shares a common meta-learner. The learner updates parameters, which are initialized by the parameters of the meta-learner, to minimize the loss on the support data. Then loss on the query data inversely instructs the training of the meta-learner. A detailed illustration of the episode-based meta-training is shown in Figure \ref{fig:episode}. During testing, only the first four operations in Figure \ref{fig:episode} are executed for a given testing episode/task. In other words, the meta-learner will not be updated during testing and the query loss becomes the testing result.

\begin{figure*}[!t]
\centering
\setlength{\abovecaptionskip}{0pt}
\begin{minipage}[t]{0.5\linewidth}
\centering
\includegraphics[width=0.85\textwidth]{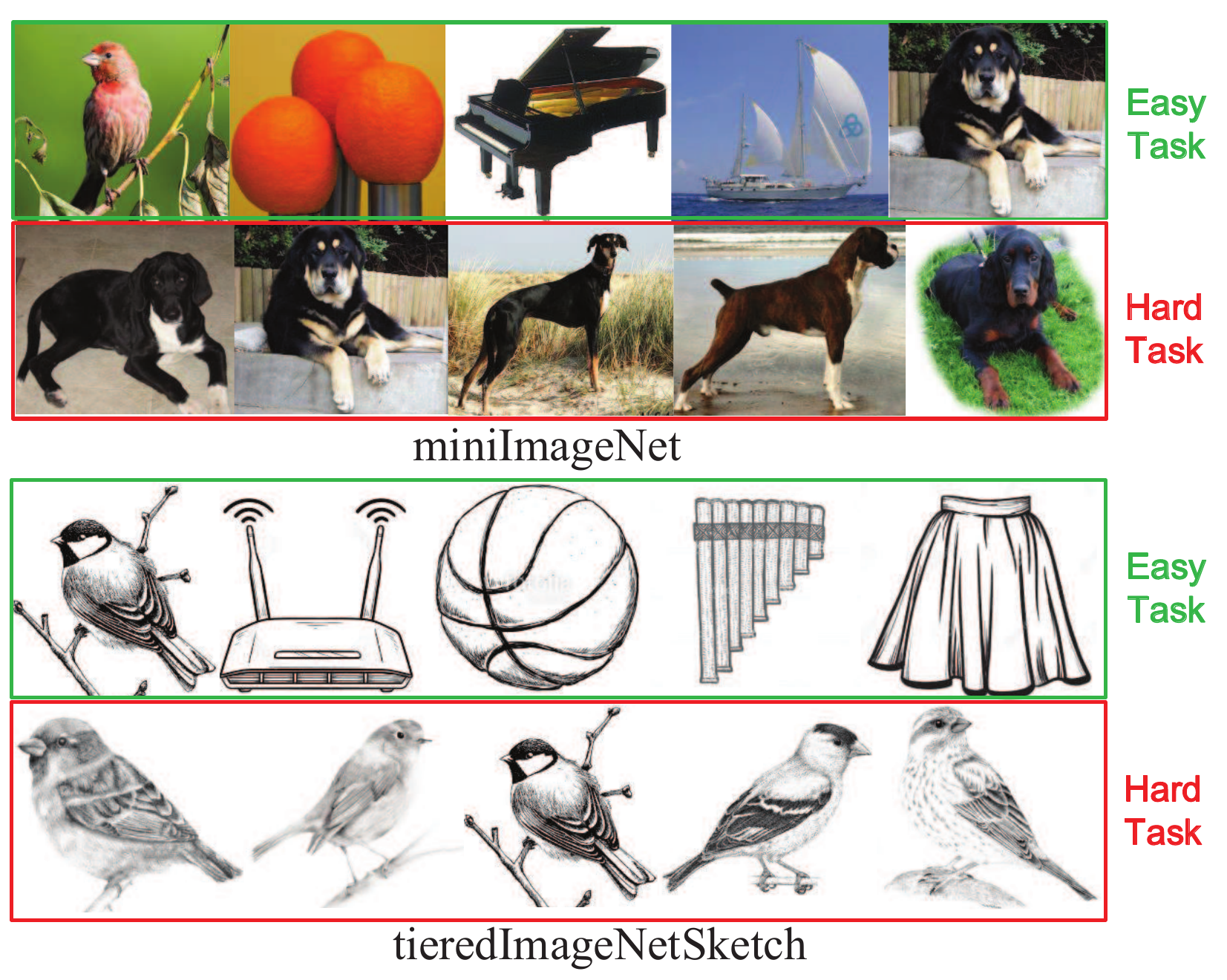}
\caption{Tasks with different hardness.}
\label{fig:hardness-episode}
\end{minipage}%
\begin{minipage}[t]{0.5\linewidth}
\centering
\includegraphics[width=0.9\textwidth]{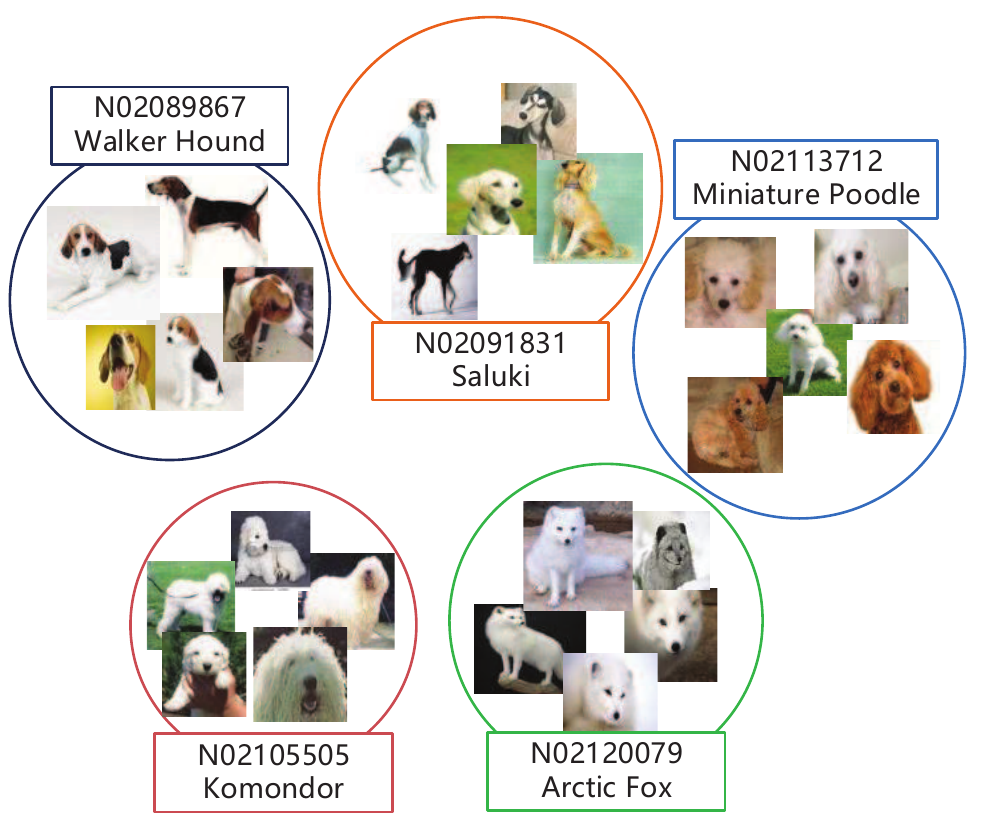}
\caption{The set-to-set relation.}
\label{fig:set-dis}
\end{minipage}%
\end{figure*}

Usually, the randomly sampled episodes are treated equally in meta-training. However, the hardness of distinguishing categories varies in different episodes. For example, Figure \ref{fig:hardness-episode} depicts two groups of tasks, where each group contains an easy task and a hard task from the same dataset. An easy task usually leads to a small loss while a hard task leads to a big one, which have different effects on the meta learner. Thus, treating them equally is not a good way. Humans learn new tasks following an efficient easy-to-hard way, which has already inspired curriculum learning to make the classifier converge faster and perform better \cite{bengio2009curriculum,shrivastava2016training}. This motivates us to consider exploring the curriculum strategy in meta-training. Note that Sun et al. \cite{sun2019meta} have already introduced the hard task concept into meta-learning. They define a hard task as the one resampled from the categories that are hard to be classified in the previous tasks. However, randomly sampling from hard categories might not obtain a hard task since the selected hard categories might be easy to differ from each other. In other words, paper \cite{sun2019meta} is more considering hard-category mining while we consider the hardness at the task level, which depends on the distinguishability among the categories in a task.

In particular, in this paper, we design an expert training strategy to obtain a more powerful meta-learner. An additional task hardness aware module (THAM) is inserted into the procedure of episode-based meta-training to evaluate the hardness of each task. We divide the original indiscriminate episode-based training procedure into two phases: a primary learning phase and an advanced training phase. In the first phase, easy tasks are preferred to make the meta-learner acquire the basic classification ability. Then, in the second phase, hard tasks are emphasized to make the primary meta-learner gradually become an expert. Such an expert training strategy is essentially to mimic the easy-to-hard learning process of human beings.

THAM is the most important component of our expert training. As the hardness of a task is highly related to the distinguishability of its categories \cite{Bengio2013Label}, which can be measured by semantic or computable relationships among its categories, we thus propose two implementations of THAM: semantic THAM (Se-THAM) and computable THAM (Co-THAM). Se-THAM directly generates easy and hard tasks based on a tree-shaped label hierarchy extracted from WordNet. It constructs a hard task with categories sharing a common superclass, while categories sampled from different superclasses make up an easy task. While training tasks in Co-THAM are still randomly sampled, Co-THAM assigns a hardness score to each task in one batch to reweight the loss according to the training phase. The harness score is a statistical measure of the relations between any two categories. Since each category contains multiple samples, the relation between any two categories is a set-to-set relation (Figure \ref{fig:set-dis}), for which we consider three types of distance measures: the pairwise Euclidean distance, the Hausdorff distance \cite{huttenlocher1993comparing}, and the Hilbert Schmidt independence criterion \cite{niu2013iterative}. Moreover, we calculate the hardness score based on features extracted from the learner, which makes Co-THAM end-to-end trainable with the meta-learner. As the meta-learner improves, the features become more reliable and the Co-THAM gets improved as well.

The contributions of this work can be summarised as follows.
\begin{itemize}
 \item We propose the easy-to-hard expert training strategy in meta-training. With this strategy, the original uniform training procedure is divided into two phases, where easy tasks and hard tasks are differently emphasized. Such strategy is easy to implement and can be applied to most meta-learning methods.
 \item We design the task hardness aware module with two implementations: Se-THAM and Co-THAM. Se-THAM samples easy and hard tasks based on the semantic relations for different learning stages. We design Co-THAM with three distance measure options, which dynamically assigns a hardness score to each randomly sampled task. To our best knowledge, this is the first work to comprehensively explore the hardness of a task in meta-learning.
 \item We conduct extensive experiments on the miniImageNet and tiereImageNetSketch datasets. The results show that the meta-learner trained with our expert training strategy and THAM generalizes better on novel tasks.
\end{itemize}
The rest of this paper is organized as follows. In Section 2, we talk about the related work. Section 3 shows the details of the proposed expert training strategy and the definitions of the hardness of a task. Experimental results and analysis on two image datasets are presented in Section 4. Finally, we conclude our work and talk about future work in Section 5.
\section{Related Work}
In this section, we review the existing works closely related to this research.
\subsection{Few-Shot Classification}
Few-shot classification aims to recognize novel categories with only a few training samples. Some researchers introduced transfer learning and semi-supervised strategies \cite{luo2017label,douze2018low}. Zhang et al. proposed a range loss and Ye et al. designed a triplet ranking loss \cite{zhang2017range} to effectively utilize the pairwise and triple relation among the categories with several samples in training.

As the main issue for the few-shot classification is that limited training samples fail to depict the intra-class diversity, many researchers devote to hallucinate training data. Hariharan et al. \cite{hariharan2017low} tried to propagate the difference of pairwise samples from a category with sufficient training data to samples of a novel category to generate new samples. Liu et al. \cite{liu2018feature} proposed an encoder-decoder based network to generate new samples with different poses. Instead of generating raw images, Xian et al. \cite{xian2018feature} employed the generative adversarial network to produce discriminative CNN features with the help of class-level semantic information. Gao et al. \cite{gao2020zero-vae-gan} coupled variational autoencoder and generative adversarial network to generate high-quality unseen features with category-level discriminability .

Besides, the meta-learning methods are also widely studied for the few-shot classification, which will be discussed next.
\subsection{Meta-Learning}
Meta-Learning, referred as learning to learn, is a popular solution for the few-shot classification. The main idea is to learn some meta-knowledge from massive related seen tasks, and then apply this meta-knowledge to quickly solve new unseen tasks. As the definition of meta-knowledge varies, the meta-learning methods can be generally divided into three categories.

\textbf{Learning an embedded feature space} supposes that all the tasks are solved by the nearest neighbor algorithm \cite{koch2015siamese,vinyals2016matching}. Snell et al. \cite{snell2017prototypical} extended this type of methods by using the center of each category in a task as the reference instead of the sample pairwise nearest neighbor. Sung et al. \cite{sung2018learning} learned this feature space by comparing the target image to a few labeled images. Different from the above methods which learn a generic feature space and then recognize a new sample by similarity calculation, Wang et al. \cite{wang2019tafe} designed a task-aware meta-learner to learn dynamic features, where a task-independent linear boundary can separate positive and negative samples. Wei et al. \cite{wei2019piecewise} tried to find sub-features responsible for recognizing some parts of a categories for few-shot fine-grained image classification.

\textbf{Learning to optimize} regards parameters in gradient-based optimization algorithms as the meta-knowledge. Ravi et al. \cite{ravi2017optimization} defined the LSTM as a meta-learner to learn the initialization and updating ways of a learner to make it quickly converge. Finn et al. \cite{finn2017model} deemed that a few fixed gradient updates for a learner can achieve good performance, and thus they designed a meta-learner for learning the weight initialization. Then, the updating strategy was further simplified by Li et al. \cite{li2017meta}, where only one-step gradient update is good enough.

\textbf{Learning to remember} introduces an augmented memory to store novel categories. Santoro et al. \cite{santoro2016one} introduced the Neural Turing Machines as an augmented memory to quickly encode and retrieve new information. Kaiser et al. \cite{kaiser2017learning} presented a large-scale life-long memory module which exploits fast nearest-neighbor algorithms for the life-long and one-shot learning.

Different from these meta-learning methods which focus on defining good meta-knowledge, our expert training acts in the training procedure and can be applied to most meta-learning methods that follow the episode-based training.
\subsection{Hardness of Samples and Label Relation}
The main idea of expert training is inspired by the effectiveness of hard samples in learning a classifier. The well-known support vector machine \cite{suykens1999least} takes the advantage of hard samples around the hyperplane to effectively promote classifier learning. Active learning \cite{balcan2007margin} and human in the loop \cite{fails2003interactive} methods first use models to select hard examples automatically and then ask humans for annotation. After that, the labeled hard examples are returned to the models for further training to improve the classification capacity.

The hardness of a task is highly related to the distinguishability of its categories, which can be generally grouped into three types in the literature. The first one is based on the high-level semantic structure (e.g., WordNet),  where categories sharing the same superclass are more alike \cite{verma2012learning,ordonez2013large,Kuang2018Integrating}.
Secondly, Griffin et al. \cite{Griffin2008Learning}, Bengio et al. \cite{Bengio2013Label} and Liu et al. \cite{Liu2013Probabilistic} regarded the classification confusion as the distinguishability of categories. They trained a classifier to obtain the confusion matrix of all categories.
Thirdly, some researchers measured the distinguishability of categories with the similarity of different categories in the feature space. Fan et al. used the Euclidian distance between the central point of each category to describe the inter-class distance \cite{Fan2015Hierarchical}. Dong et al. applied the averaged kernel distance \cite{Dong2013Training}, and Qu et al. utilized the mean and variance to measure the inter-class similarity\cite{Qu2017Joint}. Zheng et al. introduced the Hausdorff distance to describe the similarity of different sets of categories \cite{Zheng2017Hierarchical}.

Following these existing works, we investigate different task hardness measurements to evaluate the distinguishability of the categories in a task, which are well integrated into the meta-learning.
\section{The Proposed Method}
Generally, the meta-learner training works on an auxiliary dataset, which is usually organized as a dictionary $Data\_Dic = \{\{\bm{x}_j\}_{j=1}^{S_i}, C_i\}_{i=1}^M$, where $M$ is the number of categories, and $S_i$ is the number of samples in category $C_i$. This data structure makes it convenient to sample categories and the corresponding support and query examples to make up an episode.

As aforementioned, our expert training is a two-stage meta-training method. In the primary learning phase, easy tasks are required, whereas hard tasks are employed in the advanced training phase. To separate tasks, a THAM is integrated into meta-training to estimate the hardness of a task. It has two implementations. One is the Se-THAM, which constructs easy and hard tasks based on the semantic relations among the categories. The other is the Co-THAM, where the hardness of a task is a higher-order statistical estimation on the relations of its categories, which can be measured by different distance metrics. We will describe our expert training with Se-THAM and Co-THAM in the next subsections, respectively.
\subsection{Expert Training with Se-THAM}
The procedure of expert training with Se-THAM is similar to the original meta-training described in Figure \ref{fig:episode}, except that we use semantic sampling instead of the random sampling to produce easy or hard tasks for different training phases.
\begin{figure}
\setlength{\abovecaptionskip}{0pt}
\centering
\includegraphics[width=0.48\textwidth]{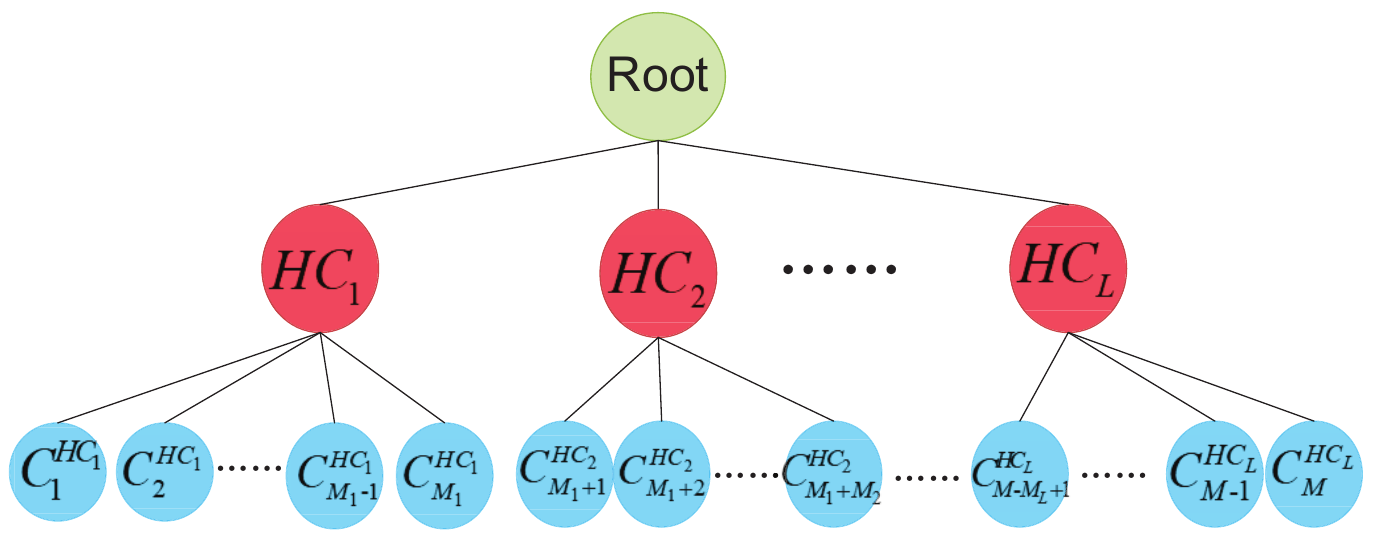}
\caption{An example of tree-shape label structure.}
\label{fig:tree}
\end{figure}

Suppose there is a tree-shaped semantic label structure for categories $\{C_i\}_{i=1}^M$ in the auxiliary dataset, and these $M$ categories can be aggregated into $L$ superclasses based on the semantic subordinate relationship (see Figure \ref{fig:tree}). We use a label dictionary to store this relation $Label\_Dic = \{HC_i, \{C_j\}_{j=1}^{M_i}\}_{i=1}^{L}$, where $M_i$ is the number of categories belonging to the superclass $HC_i$, and $M = \sum_{i=1}^LM_i$. Intuitively, categories share the same superclass are more similar than those with different superclasses. Therefore, we define that easy tasks are those whose categories come from different superclasses, and hard tasks contain categories from the same superclass. For the $N$-way-$K$-shot novel tasks, we randomly select $N$ superclasses and then choose one category from each selected superclasses to make up an easy task. While to construct a hard task, we randomly select one superclass, and then $N$ categories from this superclass. For each selected category, we sample $K$ support and $Q$ query images as an episode for the corresponding task. As the expert training with Se-THAM acts through controlling the construction of the episodes, here we just describe the sampling procedure of easy and hard tasks (in Algorithm \ref{alg:semantic-sampling}).
\begin{algorithm}[!h]
\caption{Semantic Sampling}
\label{alg:semantic-sampling}
  \textbf{Input}: Label dictionary: \\
  \qquad\quad $Label\_Dic = \{HC_i, \{C_j\}_{j=1}^{M_i}\}_{i=1}^{L}$,\\
  \qquad\quad Data Dictionary: \\
  \qquad\quad $Data\_Dic = \{\{\bm{x}_j\}_{j=1}^{S_i}, C_i\}_{i=1}^M$, \\
  \qquad\quad Task Settings: $N,K,Q$.\\
  \textbf{Output}: Target task: $\{\{\bm{x}_k^{C'_i}\}_{k=1}^K,\{\bm{x}_q^{C'_i}\}_{q=1}^Q\}_{i=1}^N$.\\
  \text{\textbf{Easy Task:}}\\
  $\{HC'_i\}_{i=1}^N = \text{random.sample}(\{HC_i\}_{i=1}^L, N)$;\\
  \For{$i=1:N$}
  {
     $C'_i = \text{random.sample}(Label\_Dic[HC'_i], 1)$;\\
  }
  \text{\textbf{Hard Task:}}\\
  shuff\_keys = random.shuffle($\{HC_i\}_{i=1}^L$);\\
  candidate = Label\_Dic[shuff\_keys[0]];\\
  \# for the situation where the selected super-class with less than N categories\\
  k = 1; \\
  \While {\text{len(candidate) }$<N$}
  {
       candidate.extend(Label\_Dic[shuff\_keys[k]]);\\
       k = k + 1;\\
  }
  $\{C'_i\}_{i=1}^N = \text{random.sample}(\text{candidate}, N)$;\\
  \text{\textbf{Select Samples:}}\\
  \For{$i=1:N$}
  {
     random.shuffle(Data\_Dic[$C'_i$]);\\
     $\{\bm{x}_k^{C'_i}\}_{k=1}^K = Data\_Dic[C'_i][1:K]$;\\
     $\{\bm{x}_q^{C'_i}\}_{q=1}^Q = Data\_Dic[C'_i][K+1:K+Q]$.\\
  }
\end{algorithm}
\subsection{Expert Training with Co-THAM}
\begin{figure*}[ht]
\setlength{\abovecaptionskip}{0pt}
\centering
\includegraphics[width=0.95\textwidth]{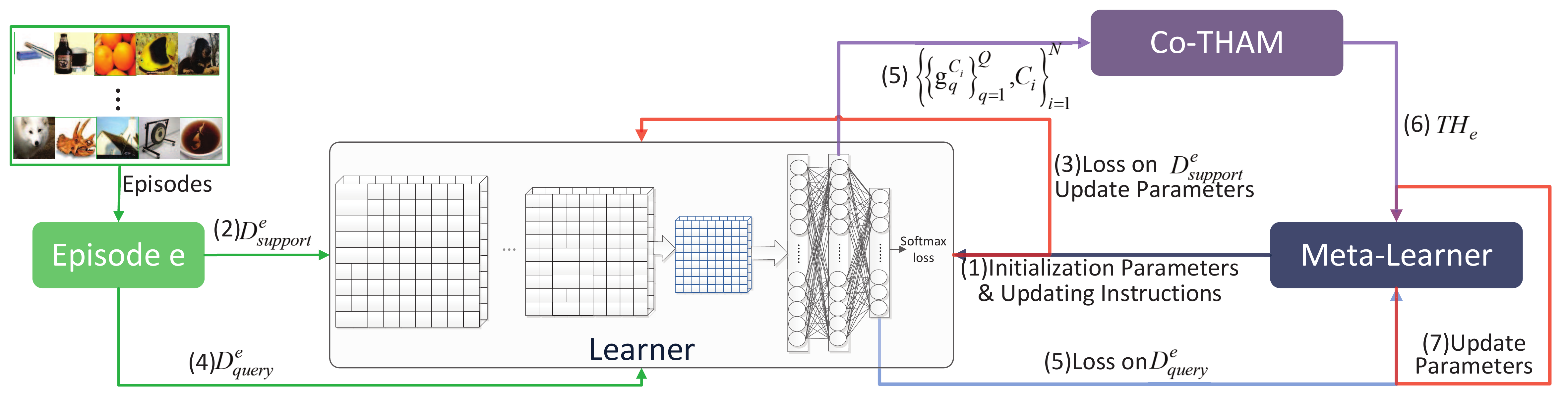}
\caption{Pipeline of the proposed expert training with Co-THAM. Operations are almost the same with those in Figure \ref{fig:episode}, expect that Co-THAM is integrated to estimate the hardness of a task with features extracted by the learner on the query data.}
\label{fig:expert-training}
\end{figure*}
\newcounter{TempEqCnt}                         
\setcounter{TempEqCnt}{\value{equation}} 
\setcounter{equation}{6}            
\begin{figure*}[hb]
\begin{equation}
\label{eq:ori-meta-loss}
\min_{\bm{\theta}}\sum_{t=1}^TL_t(f_{\bm{\theta'}_t}(\{C_i^t,\{\bm{x}_q^{C_i^t}\}_{q=1}^Q\}_{i=1}^N))
=\min_{\bm{\theta}}\sum_{t=1}^{T/B}\sum_{j=(t-1)B+1}^{tB}\frac{TH'_j}{\sum_{l=(t-1)B+1}^{tB}TH'_l}L_j(f_{\bm{\theta}-\bm{\alpha}\bigtriangledown_{\bm{\theta}}L_j(f_{\bm{\theta}})}(\{C_i^j,\{\bm{x}_q^{C_i^j}\}_{q=1}^Q\}_{i=1}^N)).
\end{equation}
\end{figure*}
Figure \ref{fig:expert-training} shows the pipeline of expert training with Co-THAM, which is integrated into the episode-based training procedure and can be trained end-to-end with the meta-learner. This expert training model contains four submodules: a randomly sampling module to produce episodes for meta-training, a learner which is usually a simple convolutional neural network (CNN) to classify a specific task, a meta-learner to produce weight initialization and updating instructions for the learners, and the Co-THAM to assign a hardness score to each task based on features extracted by the fine-tuned learner.

For a randomly sampled episode $e$, its support data $D^e_{support}=\{\{x_k^{C_i}\}_{k=1}^K,C_i\}_{C_i=1}^N$ is first used to slightly fine-tune the learner to update the initialized weight parameters from the meta-learner. Then, the fine-tuned CNN learner is applied to the query data $D^e_{query}=\{\{x_q^{C_i}\}_{q=1}^Q,C_i\}_{C_i=1}^N$ to generate two kinds of outputs. One is the loss of these query samples and the other is the features extracted from its last fully connected layer $\bm{G}_e = \{\{\bm{g}_q^{C_i}\}_{q=1}^Q, C_i\}_{i=1}^N$. Subsequently, Co-THAM uses $\bm{G}_e$ to calculate the task hardness score ${TH}_e$ and feeds it back to the meta-learner. Finally, in one mini-batch, the meta-learner receives several query losses with the corresponding task hardness scores, re-weights them with the hardness scores according to the learning stage, and then updates parameters with the re-weighted losses. In the primary learning phase, easy tasks receive higher weights, while hard tasks are given higher weights in the advanced learning phase,

As the meta-learner improves, the learner gets improved as well. Consequently, our Co-THAM also becomes more efficient with the better extracted features by the learner. With a more accurate task hardness evaluation, the meta-learner can be further enhanced greatly.
\subsubsection{Task Hardness}
Co-THAM is the most important part of our expert training, which measures the hardness of an episode/task based on the received features $\bm{G}_e = \{\{\bm{g}_q^{C_i}\}_{q=1}^Q, C_i\}_{i=1}^N$. Intuitively, the hardness of a task should be closely related to the distinguishability of its categories. For a task, if two categories are very similar (e.g., whale and shark), even though the other categories are quite different to each other (e.g., dog, desk, and church), we still treat it as a hard task as its samples cannot be easily recognized because of the two similar categories. Therefore, we define the hardness of a task as
\setcounter{equation}{0}
\begin{equation}
\label{eq:set-similarity}
{\text{TH}}_e^{Dist} = \frac {1}{\min\{Dist(C_i,C_j)\}_{i,j=1,i \neq j}^N},
\end{equation}
where $Dist(\cdot )$ denotes the distance of two categories. Essentially, Equation \eqref{eq:set-similarity} tells that the smaller the distance is, the higher the hardness score of a task or the more difficult the task will be. A straightforward distance measurement is based on the pairwise Euclidean distance over all sample pairs:
\begin{equation}
\label{eq:Euclidean}
Dist(C_i,C_j) = \sqrt{(\frac {1}{M_iM_j}{\sum_s\sum_t\|\bm{g}_s^{C_i}-\bm{g}_t^{C_j}\|^2})}.
\end{equation}
Another way to measure the distance of two sets is the Hausdorff distance, which is the maximum distance of a category to the nearest samples in the other category:
\begin{equation}
\label{eq:Hausdoroff1}
h(C_i,C_j) = \max_{\bm{g}\in C_i}\{\min_{\bm{g'} \in C_j}\{d(\bm{g}, \bm{g'})\}\},
\end{equation}
\begin{equation}
\label{eq:Hausdoroff}
H(C_i,C_j) = \max(h(C_i, C_j), h(C_j, C_i)),
\end{equation}
where $d$ can be any metric between two samples. For simplicity, here we use the Euclidian distance.

What's more, considering one category can be regarded as a random variable, the Hilbert-Schmidt independence criterion (HSIC) can also be utilized to represent the relationship of two categories. Denoting $\bm{G}_{C_i}=[\bm{g}_1^{C_i},\bm{g}_2^{C_i},\cdots, \bm{g}_Q^{C_i}]$ and $\bm{G}_{C_j}=[\bm{g}_1^{C_j},\bm{g}_2^{C_j},\cdots, \bm{g}_Q^{C_j}]$ are the feature matrices for class $C_i$ and $C_j$, the HSIC value can be computed as
\begin{equation}
\label{eq:HSIC}
\text{HSIC}(C_i, C_j) = tr(\bm{K_iHK_jH}),
\end{equation}
where $\bm{K_i}=\bm{G}_{C_i}\bm{G}_{C_i}^\mathrm{T}$ and $\bm{K_j}=\bm{G}_{C_j}\bm{G}_{C_j}^\mathrm{T}$. $\bm{H}=\bm{I}-\frac{1}{Q}\bm{1}_Q\bm{1}_Q^\mathrm{T}$, where $\bm{1}_Q\in R^Q$ is a column vector with elements of $1$. Considering two similar categories typically have a higher HSIC, we define the corresponding task hardness as:
\begin{equation}
\label{eq:HSIC-hardness}
{\text{TH}}_e^{\text{HSIC}} = \max\{\text{HSIC}(C_i,C_j)\}_{i,j=1,i \neq j}^N.
\end{equation}
\subsubsection{Training}
Parameters of the meta-learner $\bm{\theta}$ are updated by optimizing the performance of the classifier $f(\bm{\theta'})$ for each training task which are randomly sampled from the auxiliary dataset, where $\bm{\theta'}$ is initialized with $\bm{\theta}$ and updated on the support samples. For $T$ training tasks, we train the meta-learner in a mini-batch way with a batch size of $B$. In one batch, the hardness scores are adopted to re-weight the training loss in Equation \ref{eq:ori-meta-loss}. Where $TH'_j$ controls the relative importance of each task in one batch, and it takes different forms at different training phases:
\setcounter{equation}{7}
\begin{equation}
\label{easy-hard}
TH'_j =
\begin{cases}
\frac{1}{TH_j},&\text{in the Primary Learning Phase,}\\
TH_j,&\text{otherwise,}
\end{cases}
\end{equation}
where $TH_j$ is the hardness score for episode $j$ and can be calculated with Equation \eqref{eq:set-similarity} or \eqref{eq:HSIC-hardness}. Therefore, Equation \eqref{eq:ori-meta-loss} essentially gives easy tasks higher weights in the primary learning phase and assigns hard tasks higher weights in the latter phase.

For the few-shot classification, we use the cross-entropy loss for the learner and the meta-learner:
\begin{equation}
\begin{split}
\label{eq:cross-entropy-loss}
L_j(f_{\bm{\theta'}}) &= \sum_{i=1}^N\sum_{k=1}^K\mathbbm{1}(y={C_i^j})\log{f_{\bm{\theta'}}(\bm{x}_k^{C_i^j})}\\
      &+(1-\mathbbm{1}(y={C_i^j}))\log{(1-f_{\bm{\theta'}}(\bm{x}_k^{C_i^j}))},
\end{split}
\end{equation}
where $L_j$ is the loss for the $j$-th task, $\bm{x}_k^{C_i^j}$ is the k-th support sample in the $i$-th class of the $j$-th task, and $y$ is the corresponding ground-truth label.

Algorithm \ref{alg:hardness-training} gives a detailed description of the meta-training in our two-stage expert training method with Co-THAM. Function $f'_{\bm{\theta}}$ in the algorithm returns features of the last fully connected layer instead of the classification results. Note that in Algorithm \ref{alg:hardness-training}, the learner has the same network parameters $\bm{\theta}$ as the meta leaner. The learner updates its parameters from $\bm{\theta}$ to $\bm{\theta'}$ at each task, while the parameters $\bm{\theta}$ are only updated by the meta-learner at each min-batch of B tasks.
\begin{algorithm}[!t]
\caption{Expert Training with Co-THAM}
\label{alg:hardness-training}
  \textbf{Input}: The auxiliary dataset: $D=\{\{\bm{x}_j\}_{j=1}^{S_i}, C_i\}_{i=1}^M$, \\
  \qquad\quad Updating hyperparameters: $\bm{\alpha}, \bm{\beta}$.\\
  \qquad\quad Task Settings: $N,K,Q$.\\
  \textbf{Output}: $\bm{\theta}$
  $\{\{\{\bm{x}_k^{C_i^t}\}_{k=1}^K,\{\bm{x}_q^{C_i^t}\}_{q=1}^Q\}_{i=1}^N\}_{t=1}^T$=task.sample($D$);\\
  Randomly initialize $\bm{\theta}$;\\
  \For{$t=1:T/B$}
  {
       \For{$j=(t-1)B+1:tB$}
       {
            Calculate $\bigtriangledown_{\bm{\theta}}L_j(f_{\bm{\theta}})$ with $\{\{\bm{x}_k^{C_i^j}\}_{k=1}^K\}_{i=1}^N$ and $L_j$ in Equation (\ref{eq:cross-entropy-loss});\\
            Updata $\bm{\theta}$ with gradient descent:
            $\bm{\theta'} = \bm{\theta}-\bm{\alpha}\bigtriangledown_{\bm{\theta}}L_j(f_{\bm{\theta}})$;\\
            Extract features for the query data:\\
            $\{\{\bm{g}_q^{C_i^j}\}_{q=1}^Q\}_{i=1}^N = f'_{\bm{\theta'}}(\{\{\bm{x}_q^{C_i^j}\}_{q=1}^Q\}_{i=1}^N)$;\\
            Evaluate the task hardness with Equation (\ref{eq:set-similarity}) or (\ref{eq:HSIC-hardness}):\\
            $\text{THS}[j] = TH_j(\{\{\bm{g}_q^{C_i^j}\}_{q=1}^Q, C_i^j\}_{i=1}^N)$;\\
            Compute the meta-loss with Equation (\ref{eq:cross-entropy-loss}):
            $L_j(f_{\bm{\theta'}})=L_j(f_{\bm{\theta'}}(\{\{\bm{x}_q^{C_i^j}\}_{q=1}^Q, C_i^j\}_{i=1}^N)))$;\\
            LS$[j]=L_j(f_{\bm{\theta'}})$;\\
       }
       \if{in Primary Learning}
       {
            THS = $\frac{1}{THS}$;\\
       }
       \fi
       Re-Weight the meta-loss in one batch:\\
       $L_t(f_{\bm{\theta'}}) = \frac{\sum_{j=(t-1)B+1}^{tB}\text{THS}[j]LS[j]}{\sum_{j=(t-1)B+1}^{tB}\text{THS}[j]}$;\\
       Update the parameter:
       $\bm{\theta} = \bm{\theta}-\bm{\beta}\bigtriangledown_{\bm{\theta}}L_t(f_{\bm{\theta'}})$;\\
  }
\end{algorithm}

\section{Experiment}
\subsection{Datasets}
We conduct our experiments on two datasets: miniImageNet and tieredImageNetSketch.

The miniImageNet dataset \cite{vinyals2016matching} contains 60,000 colour images belonging to 100 classes. 64 of these classes are used for training a meta-learner, 16 classes for validating, and the remaining 20 classes for testing. Since it is a subset of ImageNet, we can directly extract the label structure for the semantic sampling from that of ImageNet, which organizes all the classes from coarse-grained to fine-grained levels. For the semantic sampling, we split the label structure for the training classes from the extracted label structure.

The tieredImageNetSketch dataset is constructed with the images in ImgenetSketch \cite{wang2019learning} and the label structure from tieredImageNet \cite{ren2018meta}. It consists of 50,000 sketch images, 50 images for each of the 1000 classes in ImageNet. As it borrows the label structure from tieredImageNet, categories in tieredImageNetSketch are the same as tieredImageNet. It contains 608 classes which are grouped into 34 superclasses. These superclasses are further split into 20 for training, 6 for validation and 8 for testing, which ensures that the training classes are sufficiently distinct from the testing classes. The reason we use tieredImageNetSketch instead of tieredImageNet is that we want to make the datasets more diverse, as both tieredImageNet and miniImageNet are subsets of ImageNet.

During meta-training, samples of the training classes make up the auxiliary dataset to generate $T$ training tasks. During testing, $V$ novel tasks are randomly sampled from the testing classes, where the $N\times Q$ query samples in each novel task are used to calculate the classification accuracy. Then, the mean and the standard deviation of the classification accuracies over the $V$ novel tasks are utilized to measure the performance. We introduce $\lambda$ to set when to start the second training phase, where the first $\lambda\times T$ tasks are responsible for the primary learning and the remaining tasks are for the second stage training.

\subsection{Model Study}
There are two variables in our expert training: THAMs and $\lambda$. To study their effects, we train Meta-SGD \cite{li2017meta} on miniImageNet. For a fair comparison, the network architecture and learning rates for the learner and the meta-learner are the same as those in~\cite{li2017meta}. We set $T=60,000$, $V=600$, and $Q=20$.

Table \ref{tab:different-hardness} shows the results with $\lambda=1/4$ for the 5-way-1-shot task and $\lambda=1/3$ for the 5-way-5-shot task. In the table, Meta-SGD-our means the Meta-SGD is trained on our re-sampled dataset. Meta-SGD-ET-Se-THAM stands for the Meta-SGD trained with our expert training and Se-THAM. Meta-SGD-ET-Co-THAM-Pairwise, Meta-SGD-ET-Co-THAM-Hausdorff, and Meta-SGD-ET-Co-THAM-HSIC represent that the inter-class relations are estimated with the pairwise Euclidean distance, the Hausdorff distance, and the HSIC, respectively.
\begin{table}[!htb]
\centering
\caption{Results with Different THAMs on miniImageNet.}
\begin{tabular}{|c|c|c|}
\hline
\multirow{2}{*}{Definition of Hardness}
&\multicolumn{2}{|c|}{5-way Accuracy}\\
\cline{2-3}
&1-shot&5-shot\\
\hline
Meta-SGD-$our$ &49.50$\pm$1.78\%&63.90$\pm$1.58\% \\
\hline
Meta-SGD-ET-Se-THAM  & 48.20$\pm$2.01\% & \textbf{64.64$\pm$1.37\%}    \\
\hline
Meta-SGD-ET-Co-THAM-Pairwise  & \textbf{50.64$\pm$1.83\%}  & \textbf{64.81$\pm$1.49\%}    \\
\hline
Meta-SGD-ET-Co-THAM-Hausdorff  & \textbf{49.86$\pm$1.68\%}  & \textbf{67.42$\pm$1.43\%}   \\
\hline
Meta-SGD-ET-Co-THAM-HSIC  &\textbf{50.58$\pm$1.70\%}  & \textbf{67.42$\pm$1.73\%}   \\
\hline
\end{tabular}
\label{tab:different-hardness}
\end{table}
\begin{figure*}[!hbt]
\setlength{\abovecaptionskip}{0pt}
\centering
\includegraphics[width=0.8\textwidth]{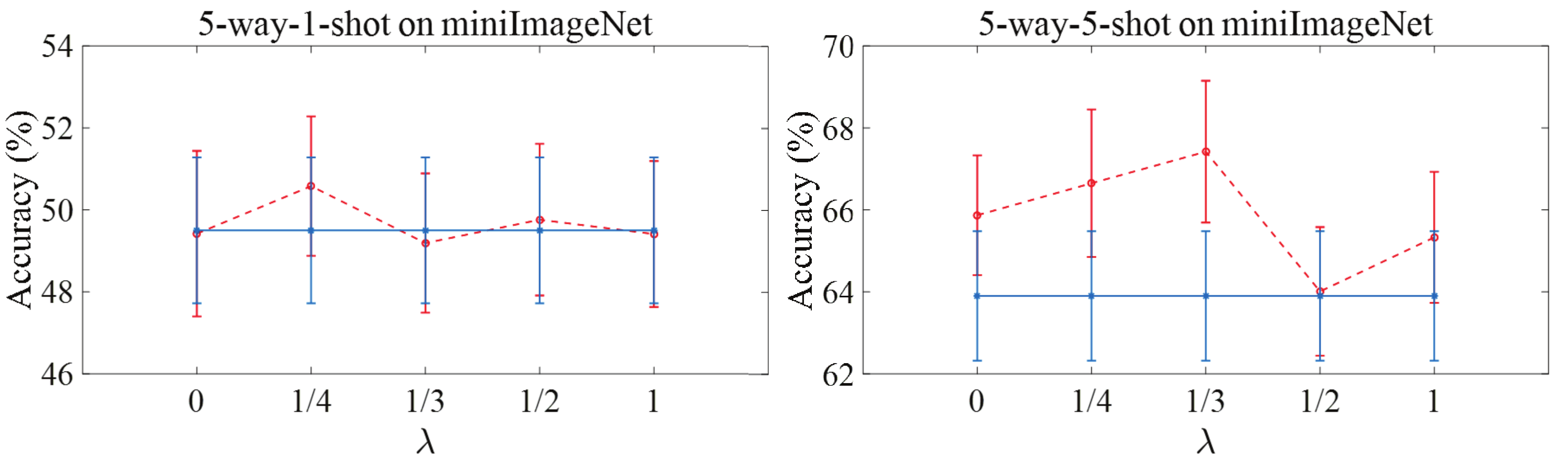}
\caption{Accuracy with different proportions of easy tasks in expert training. The blue line shows the results of Meta-SGD, and the red dashed line represents those of Meta-SGD-ET-Co-THAM-HSIC}
\label{fig:partnum}
\end{figure*}

The results demonstrate that: (1) Meta-SGD-ET-Co-THAM-HSIC achieves the best results on both 5-way-1-shot tasks and 5-way-5-shot tasks. (2) Meta-SGD-ET-Se-THAM performs worse than the expert training with Co-THAM, because tasks sampled by Se-THAM are strictly easy or hard. While Co-THAM is `softer' by assigning a continuous hardness score to each task and making all the tasks contribute to training the meta-learner. (3) For 5-way-5-shot tasks, Co-THAM with the Hausdorff distance and HSIC are better than that with the pairwise Euclidean distance, which suggests the benefit of considering the variations in a category.

In the following experiments, we choose Co-THAM-HSIC to compare with other models for four reasons. Firstly, HSIC achieves consistently better results on two types of tasks. Secondly, Hausdorff is theoretically simple, but it is computationally difficult because of the densely minimum sample pairwise distance estimation. Thirdly, HSIC regards each category as a distribution, and all the samples can contribute to the relation estimation. While, Hausdorff measures the distance of two categories with the maximum nearest sample pair, making it more sensitive to outliers. Finally, as training samples are limited, models are always shallow. Thus, the extracted features are not powerful enough to measure relations in the linear Euclidean space. HSIC can benefit from mapping the features into a non-linear kernel space.

Figure \ref{fig:partnum} describes the results with $\lambda$ varying in [0, 1/4, 1/3, 1/2, 1]. The blue line is the results of Meta-SGD, and the red dashed line represents Meta-SGD-ET-Co-THAM-HSIC. We can see that 5-way-5-shot tasks require more easy tasks than 5-way-1-shot tasks to obtain better results. Thus, we set $\lambda=1/4$ for 5-way-1-shot tasks and $\lambda=1/3$ for 5-way-5-shot tasks in the following.

\subsection{Experimental Results and Analysis}
We implement our expert training with Co-THAM on Meta-SGD \cite{li2017meta}. The classification results on miniImageNet and tieredImageNetSketch are shown in Table \ref{tab:results}. The results of Meta-SGD-$paper$ are obtained directly from the original paper. Meta-SGD-$our$ represents the Meta-SGD trained on our datasets with more complicated training-test partitions, so its results are slightly worse than those in Meta-SGD-$paper$. It can be concluded that Meta-SGD-ET-Co-THAM-HSIC achieves the best results on both miniImageNet and tieredImageNetSketch. The improvement in 5-way-5-shot tasks is more significant than that in 5-way-1-shot tasks. The reason may be that learners trained on 5-way-5-shot tasks are more accurate, which leads to more reliable hardness evaluations and consequently the meta-learner is fine-tuned more properly.
\begin{table}[!htb]
\centering
\caption{Results on miniImageNet (top) and tieredImageNetSketch (bottom).}
\begin{tabular}{|c|c|c|}
\hline
\multirow{2}{*}{Method}
&\multicolumn{2}{|c|}{5-way Accuracy}\\
\cline{2-3}
&1-shot&5-shot\\
\hline
\hline
  Meta-SGD-$paper$&50.47$\pm$1.87\%&64.03$\pm$0.94\%\\
  Meta-SGD-$our$&49.50$\pm$1.78\%&63.90$\pm$1.58\% \\
  Meta-SGD-ET-Co-THAM-HSIC&\textbf{50.58$\pm$1.70\%}&\textbf{67.42$\pm$1.73\%}\\
\hline
\hline
  Meta-SGD-$paper$&--&--\\
  Meta-SGD-$our$&51.24$\pm$2.04\%&67.46$\pm$1.59\%\\
  Meta-SGD-ET-Co-THAM-HSIC&\textbf{51.92$\pm$2.15\%}&\textbf{68.31$\pm$1.60\%}\\
\hline
\end{tabular}
\label{tab:results}
\end{table}

To show the effectiveness of our expert training, we train the Meta-SGD with another two training schedules, i.e., the reversed phases and the probabilistic expert training. The reversed phases reverse the two phases in expert training, where hard tasks are emphasize firstly, then easy tasks are required. In the probabilistic ET, the target tasks (easy/hard) are sampled with a higher probability of 0.8 for different training phases. Results on miniImageNet are shown in Table \ref{tab:schedules}. We can see that compared with Meta-SGD, the probabilistic ET performs worse for 5-way 1-shot tasks, while better for 5-way 5-shot tasks. The results of reversed phases show similar trend. Overall, our expert training achieves the best performance.
\begin{table}[!htb]
\centering
\caption{Results of Meta-SGD with different training schedules on miniImageNet.}
\begin{tabular}{|c|c|c|}
\hline
\multirow{2}{*}{Method}
&\multicolumn{2}{|c|}{5-way Accuracy}\\
\cline{2-3}
&1-shot&5-shot\\
\hline
\hline
  Meta-SGD-$paper$ &50.47$\pm$1.87\%&64.03$\pm$0.94\%\\
  Meta-SGD-$our$ &49.50$\pm$1.78\%&63.90$\pm$1.58\% \\
  Meta-SGD-Reversed Phases &49.29$\pm$1.60\%	&65.29$\pm$1.41\%\\
  Meta-SGD-Probabilistic-ET &47.96$\pm$1.92\%&65.60$\pm$1.54\%\\
  Meta-SGD-ET-Co-THAM-HSIC &\textbf{50.58$\pm$1.70\%}&\textbf{67.42$\pm$1.73\%}\\
\hline
\end{tabular}
\label{tab:schedules}
\end{table}

To show whether our expert training works well on other meta-learning models, we implement it on another two popular methods: MAML \cite{finn2017model} and Relation net \cite{sung2018learning}. We use the same network architectures and the learning rates as those in the papers. We set $T=60,000$, $V=500$, and $Q=20$ for MAML. For Relation net, we set $T=300,000$, $V=100$, and $Q=15$ for 5-way-1-shot tasks, and $Q=10$ for 5-way-5-shot tasks, the same as  \cite{sung2018learning}. The results in Table \ref{tab:various-methods} demonstrate that when trained with our expert training strategy, meta-learners can always perform better.
\begin{table}[!htb]
\centering
\caption{Results of Existing Methods on miniImageNet.}
\begin{tabular}{|c|c|c|}
\hline
\multirow{2}{*}{Method}
&\multicolumn{2}{|c|}{5-way Accuracy}\\
\cline{2-3}
&1-shot&5-shot\\
\hline
\hline
  MAML-$paper$&48.70$\pm$1.84\% &63.11$\pm$0.92\% \\
  MAML-$our$&48.47$\pm$2.04\%&62.98$\pm$1.65\% \\
  MAML-ET-Co-THAM-HSIC&\textbf{49.54$\pm$1.95\%}&\textbf{66.10$\pm$1.62\%} \\
\hline
\hline
  Relation net-$paper$&50.44$\pm$0.82\%&65.32$\pm$0.70\% \\
  Relation net-$our$&51.02$\pm$0.84\%& 64.78$\pm$0.68\% \\
  Relation net-ET-Co-THAM-HSIC&\textbf{51.24$\pm$0.85\%}&\textbf{66.00$\pm$0.77\%} \\
\hline
\end{tabular}
\label{tab:various-methods}
\end{table}

To show the effectiveness of our expert training strategy, we compare it with meta-transfer learning (MTL), which propose to train meta-learner with hard tasks (HTs). In MTL, a hard task is made up of categories with the highest misclassification rate in each former task, i.e., hard categories in former tasks makes up the current HTs. Experimental results on MAML with HTs and our expert training are shown in Table \ref{tab:MTL}. Our MAML-ET-Co-THAM-HSIC method performs better than MAML and MAML-HT, showing the benefit of a reasonable task hardness estimation and curriculum learning.

\begin{figure}[h]
\setlength{\abovecaptionskip}{0pt}
\centering
\includegraphics[width=0.5\textwidth]{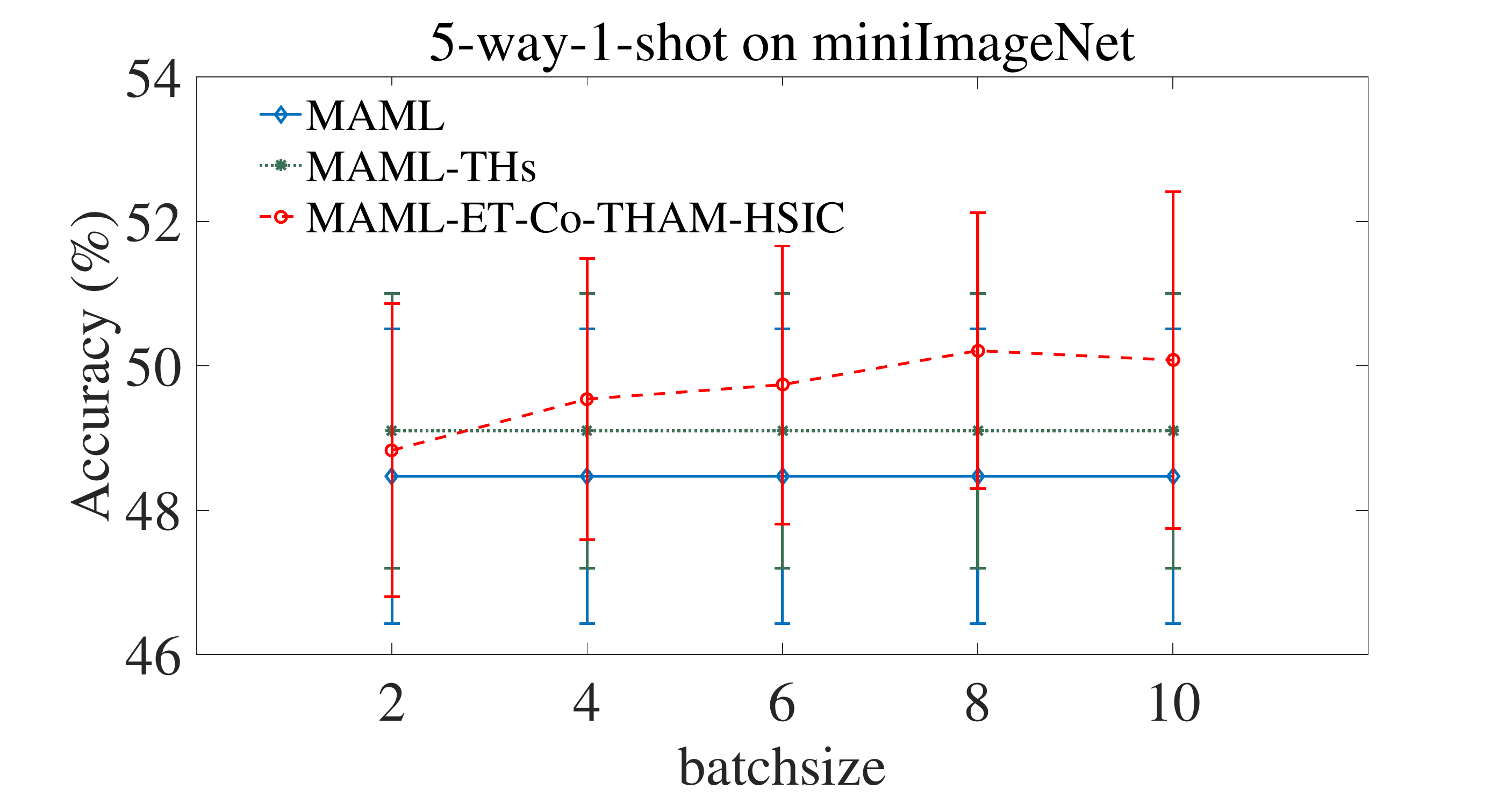}
\caption{Results of MAML trained with different strategies.}
\label{fig:batchsize}
\end{figure}

To achieve state of the art results, we also implement our expert training on MTL with a pre-trained ResNet-$12$ as the backbone. As our expert training strategy requires to retain models for each training task in a batch for reweighting the meta-loss and the ResNet12 is GPU memory consuming, the batchsize is set to 2 on the GeForce RTX 2080 with a memory of 11G. Nevertheless, our MTL-ResNet-ET-Co-THAM-HSIC still achieves better results on the 5-way-1-shot tasks, and comparable results on the 5-way-5-shot tasks (see Table \ref{tab:MTL}). Then, we vary the batchsize in [2,4,6,8,10] on a small model MAML. Results in Figure \ref{fig:batchsize} show that MAML-ET-Co-THAM-HSIC performs better with a larger batchsize. Thus, we can infer that the results of MTL-ResNet-ET-Co-THAM-HSIC can be further improved with more training tasks provided to calculate the meta-loss in a batch.
\begin{table}[!htb]
\centering
\caption{Comparison with State of the Art Methods on miniImageNet.}
\begin{tabular}{|c|c|c|}
\hline
\multirow{2}{*}{Method}
&\multicolumn{2}{|c|}{5-way Accuracy}\\
\cline{2-3}
&1-shot&5-shot\\
\hline
\hline
  MAML-$our$&48.47$\pm$2.04\%&62.98$\pm$1.65\% \\
  MAML-THs&49.10$\pm$1.90\%&64.10$\pm$0.90\% \\
  MAML-ET-Co-THAM-HSIC&\textbf{49.54$\pm$1.95\%}&\textbf{66.10$\pm$1.62\%} \\
\hline
\hline
  MTL-ResNet&60.20$\pm$1.80\%&74.30$\pm$0.90\% \\
  MTL-ResNet-THs&61.20$\pm$1.80\%&\textbf{75.50$\pm$0.80\%} \\
  MTL-ResNet-ET-Co-THAM-HSIC&\textbf{62.01$\pm$1.13\%}&\underline{75.27$\pm$0.61\%} \\
\hline
\end{tabular}
\label{tab:MTL}
\end{table}
\subsection{Results on  Testing Tasks with Different Hardness}
Usually, tasks for testing are randomly sampled, but it is interesting to see the performance on easy and hard tasks separately. For this, we introduce the semantic sampling in Algorithm \ref{alg:semantic-sampling} to generate easy and hard tasks for testing. Experimental results of the 5-way-5-shot tasks on tieredImageNetSketch are depicted in Table \ref{tab:different-hardness}. We choose tieredImageNetSketch because it contains rich testing categories to generate various easy and hard tasks. Random means that testing tasks are randomly sampled. ALL Easy ( Hard) denotes that tasks are semantically easy (hard). We can see that the improvements on ALL Easy and ALL Hard are more obvious than Random.
\begin{table}[!htb]
\centering
\caption{Results on Different Hardness of Testing Tasks.}
\begin{tabular}{|c|c|c|}
\hline
Method & Meta-SGD  &Meta-SGD-E-Co-THAM-HSIC\\
\hline
Random & 67.46$\pm$1.59\% &\textbf{68.31$\pm$1.60\%} \\
\hline
ALL Easy &81.92$\pm$1.672\% &\textbf{84.70$\pm$1.47\%} \\
\hline
ALL Hard &58.59$\pm$2.72\% &\textbf{63.02$\pm$2.65\%} \\
\hline
\end{tabular}
\label{tab:different-test}
\end{table}

\section{Conclusion}
In this paper, we have proposed an easy-to-hard expert training strategy for meta-learning, which trains easy tasks firstly followed by hard ones. We have particularly proposed a task hardness aware module (THAM) with two implementations: the semantic THAM (Se-THAM) and the computable THAM (Co-THAM). Experimental results on two image datasets have demonstrated that the meta-learners trained with the proposed expert training generalize better on novel tasks. Overall, the expert training with Co-THAM achieves better results than that with Se-THAM. We believe this is because the computable hardness is integrated into the end-to-end training procedure and is optimized as the meta-learner improves.

In the future, we will explore what the performance would be with more than two phases. Besides, our expert training strategy is GPU memory consuming for retaining multiple models in a batch, so we will find more efficient ways to implement curriculum learning in meta-learning.

\ifCLASSOPTIONcaptionsoff
  \newpage
\fi

\bibliographystyle{IEEEtran}
\bibliography{et}

\begin{thebibliography}{10}
\providecommand{\url}[1]{#1}
\csname url@samestyle\endcsname
\providecommand{\newblock}{\relax}
\providecommand{\bibinfo}[2]{#2}
\providecommand{\BIBentrySTDinterwordspacing}{\spaceskip=0pt\relax}
\providecommand{\BIBentryALTinterwordstretchfactor}{4}
\providecommand{\BIBentryALTinterwordspacing}{\spaceskip=\fontdimen2\font plus
\BIBentryALTinterwordstretchfactor\fontdimen3\font minus
  \fontdimen4\font\relax}
\providecommand{\BIBforeignlanguage}[2]{{%
\expandafter\ifx\csname l@#1\endcsname\relax
\typeout{** WARNING: IEEEtran.bst: No hyphenation pattern has been}%
\typeout{** loaded for the language `#1'. Using the pattern for}%
\typeout{** the default language instead.}%
\else
\language=\csname l@#1\endcsname
\fi
#2}}
\providecommand{\BIBdecl}{\relax}
\BIBdecl

\bibitem{deng2009imagenet}
J.~Deng, W.~Dong, R.~Socher, L.-J. Li, K.~Li, and L.~Fei-Fei, ``Imagenet: A
  large-scale hierarchical image database,'' in \emph{Proceedings of the IEEE
  Conference on Computer Vision and Pattern Recognition, CVPR}.\hskip 1em plus
  0.5em minus 0.4em\relax IEEE, 2009, pp. 248--255.

\bibitem{taigman2014deepface}
Y.~Taigman, M.~Yang, M.~Ranzato, and L.~Wolf, ``Deepface: Closing the gap to
  human-level performance in face verification,'' in \emph{Proceedings of the
  IEEE Conference on Computer Vision and Pattern Recognition, CVPR}, 2014, pp.
  1701--1708.

\bibitem{miech2019howto100m}
A.~Miech, D.~Zhukov, J.-B. Alayrac, M.~Tapaswi, I.~Laptev, and J.~Sivic,
  ``Howto100m: Learning a text-video embedding by watching hundred million
  narrated video clips,'' in \emph{Proceedings of the IEEE International
  Conference on Computer Vision, ICCV}.\hskip 1em plus 0.5em minus 0.4em\relax
  Ieee, 2019.

\bibitem{lin2014microsoft}
T.-Y. Lin, M.~Maire, S.~Belongie, J.~Hays, P.~Perona, D.~Ramanan,
  P.~Doll{\'a}r, and C.~L. Zitnick, ``Microsoft coco: Common objects in
  context,'' in \emph{European Conference on Computer Vision, ECCV}.\hskip 1em
  plus 0.5em minus 0.4em\relax Springer, 2014, pp. 740--755.

\bibitem{mondal2018few}
A.~K. Mondal, J.~Dolz, and C.~Desrosiers, ``Few-shot 3d multi-modal medical
  image segmentation using generative adversarial learning,'' \emph{arXiv
  preprint arXiv:1810.12241}, 2018.

\bibitem{tanner2018generative}
C.~Tanner, F.~Ozdemir, R.~Profanter, V.~Vishnevsky, E.~Konukoglu, and
  O.~Goksel, ``Generative adversarial networks for mr-ct deformable image
  registration,'' \emph{arXiv preprint arXiv:1807.07349}, 2018.

\bibitem{yuan2019hyperspectral}
Q.~Yuan, Q.~Zhang, J.~Li, H.~Shen, and L.~Zhang, ``Hyperspectral image
  denoising employing a spatial¨cspectral deep residual convolutional neural
  network,'' \emph{IEEE Transactions on Geoscience and Remote Sensing},
  vol.~57, no.~2, pp. 1205--1218, 2019.

\bibitem{hariharan2016low}
B.~Hariharan and R.~Girshick, ``Low-shot visual object recognition,'' 2016.

\bibitem{li2019large}
A.~Li, T.~Luo, Z.~Lu, T.~Xiang, and L.~Wang, ``Large-scale few-shot learning:
  Knowledge transfer with class hierarchy,'' in \emph{Proceedings of the IEEE
  Conference on Computer Vision and Pattern Recognition, CVPR}, 2019, pp.
  7212--7220.

\bibitem{sung2018learningToLearn}
F.~Sung, L.~Zhang, T.~Xiang, T.~Hospedales, and Y.~Yang, ``Learning to learn:
  Meta-critic networks for sample efficient learning,'' in \emph{Proceedings of
  the Cnternational Conference on Learning Representations, ICLR}, 2018.

\bibitem{ravi2017optimization}
S.~Ravi and H.~Larochelle, ``Optimization as a model for few-shot learning,''
  in \emph{Proceedings of the Cnternational Conference on Learning
  Representations, ICLR}, 2017.

\bibitem{sung2018learning}
F.~Sung, Y.~Yang, L.~Zhang, T.~Xiang, P.~H. Torr, and T.~M. Hospedales,
  ``Learning to compare: Relation network for few-shot learning,'' in
  \emph{Proceedings of the IEEE Conference on Computer Vision and Pattern
  Recognition, CVPR}, 2018, pp. 1199--1208.

\bibitem{li2017meta}
Z.~Li, F.~Zhou, F.~Chen, and H.~Li, ``Meta-sgd: Learning to learn quickly for
  few-shot learning,'' \emph{arXiv preprint arXiv:1707.09835}, 2017.

\bibitem{finn2017model}
C.~Finn, P.~Abbeel, and S.~Levine, ``Model-agnostic meta-learning for fast
  adaptation of deep networks,'' in \emph{Proceedings of the 34th International
  Conference on Machine Learning, ICML}.\hskip 1em plus 0.5em minus 0.4em\relax
  JMLR. org, 2017, pp. 1126--1135.

\bibitem{bengio2009curriculum}
Y.~Bengio, J.~Louradour, R.~Collobert, and J.~Weston, ``Curriculum learning,''
  in \emph{Proceedings of the 26th Annual International Conference on Machine
  Learning, ICML}, 2009, pp. 41--48.

\bibitem{shrivastava2016training}
A.~Shrivastava, A.~Gupta, and R.~Girshick, ``Training region-based object
  detectors with online hard example mining,'' in \emph{Proceedings of the IEEE
  Conference on Computer Vision and Pattern Recognition, CVPR}, 2016, pp.
  761--769.

\bibitem{sun2019meta}
Q.~Sun, Y.~Liu, T.-S. Chua, and B.~Schiele, ``Meta-transfer learning for
  few-shot learning,'' in \emph{Proceedings of the IEEE Conference on Computer
  Vision and Pattern Recognition, CVPR}, 2019, pp. 403--412.

\bibitem{Bengio2013Label}
S.~Bengio, J.~Weston, and D.~Grangier, ``Label embedding trees for large
  multi-class tasks,'' \emph{In Advances in Neural Information Processing
  Systems, NIPS}, pp. 163--171, 2013.

\bibitem{huttenlocher1993comparing}
D.~P. Huttenlocher, G.~A. Klanderman, and W.~J. Rucklidge, ``Comparing images
  using the hausdorff distance,'' \emph{IEEE Transactions on Pattern Analysis
  and Machine Intelligence, TPAMI}, vol.~15, no.~9, pp. 850--863, 1993.

\bibitem{niu2013iterative}
D.~Niu, J.~G. Dy, and M.~I. Jordan, ``Iterative discovery of multiple
  alternativeclustering views,'' \emph{IEEE Transactions on Pattern Analysis
  and Machine Intelligence, TPAMI}, vol.~36, no.~7, pp. 1340--1353, 2013.

\bibitem{luo2017label}
Z.~Luo, Y.~Zou, J.~Hoffman, and L.~F. Fei-Fei, ``Label efficient learning of
  transferable representations acrosss domains and tasks,'' in \emph{Advances
  in Neural Information Processing Systems, NIPS}, 2017, pp. 165--177.

\bibitem{douze2018low}
M.~Douze, A.~Szlam, B.~Hariharan, and H.~J{\'e}gou, ``Low-shot learning with
  large-scale diffusion,'' in \emph{Proceedings of the IEEE Conference on
  Computer Vision and Pattern Recognition, CVPR}, 2018, pp. 3349--3358.

\bibitem{zhang2017range}
X.~Zhang, Z.~Fang, Y.~Wen, Z.~Li, and Y.~Qiao, ``Range loss for deep face
  recognition with long-tailed training data,'' in \emph{Proceedings of the
  IEEE International Conference on Computer Vision, ICCV}, 2017, pp.
  5409--5418.

\bibitem{hariharan2017low}
B.~Hariharan and R.~Girshick, ``Low-shot visual recognition by shrinking and
  hallucinating features,'' in \emph{Proceedings of the IEEE International
  Conference on Computer Vision, ICCV}, 2017, pp. 3018--3027.

\bibitem{liu2018feature}
B.~Liu, X.~Wang, M.~Dixit, R.~Kwitt, and N.~Vasconcelos, ``Feature space
  transfer for data augmentation,'' in \emph{Proceedings of the IEEE Conference
  on Computer Vision and Pattern Recognition, CVPR}, 2018, pp. 9090--9098.

\bibitem{xian2018feature}
Y.~Xian, T.~Lorenz, B.~Schiele, and Z.~Akata, ``Feature generating networks for
  zero-shot learning,'' in \emph{Proceedings of the IEEE Conference on Computer
  Vision and Pattern Recognition, CVPR}, 2018, pp. 5542--5551.

\bibitem{gao2020zero-vae-gan}
R.~Gao, X.~Hou, J.~Qin, J.~Chen, L.~Liu, F.~Zhu, Z.~Zhang, and L.~Shao,
  ``Zero-vae-gan: Generating unseen features for generalized and transductive
  zero-shot learning,'' \emph{IEEE Transactions on Image Processing, TIP},
  vol.~29, pp. 3665--3680, 2020.

\bibitem{koch2015siamese}
G.~Koch, R.~Zemel, and R.~Salakhutdinov, ``Siamese neural networks for one-shot
  image recognition,'' in \emph{ICML deep learning workshop}, vol.~2, 2015.

\bibitem{vinyals2016matching}
O.~Vinyals, C.~Blundell, T.~Lillicrap, D.~Wierstra \emph{et~al.}, ``Matching
  networks for one shot learning,'' in \emph{Advances in neural information
  processing systems, NIPS}, 2016, pp. 3630--3638.

\bibitem{snell2017prototypical}
J.~Snell, K.~Swersky, and R.~Zemel, ``Prototypical networks for few-shot
  learning,'' in \emph{Advances in Neural Information Processing Systems,
  NIPS}, 2017, pp. 4077--4087.

\bibitem{wang2019tafe}
X.~Wang, F.~Yu, R.~Wang, T.~Darrell, and J.~E. Gonzalez, ``Tafe-net: Task-aware
  feature embeddings for low shot learning,'' in \emph{Proceedings of the IEEE
  Conference on Computer Vision and Pattern Recognition, CVPR}, 2019, pp.
  1831--1840.

\bibitem{wei2019piecewise}
X.~Wei, P.~Wang, L.~Liu, C.~Shen, and J.~Wu, ``Piecewise classifier mappings:
  Learning fine-grained learners for novel categories with few examples,''
  \emph{IEEE Transactions on Image Processing, TIP}, vol.~28, no.~12, pp.
  6116--6125, 2019.

\bibitem{santoro2016one}
A.~Santoro, S.~Bartunov, M.~Botvinick, D.~Wierstra, and T.~Lillicrap,
  ``One-shot learning with memory-augmented neural networks,'' \emph{arXiv
  preprint arXiv:1605.06065}, 2016.

\bibitem{kaiser2017learning}
{\L}.~Kaiser, O.~Nachum, A.~Roy, and S.~Bengio, ``Learning to remember rare
  events,'' in \emph{Proceedings of the Cnternational Conference on Learning
  Representations, ICLR}, 2017.

\bibitem{suykens1999least}
J.~A. Suykens and J.~Vandewalle, ``Least squares support vector machine
  classifiers,'' \emph{Neural Processing Letters}, vol.~9, no.~3, pp. 293--300,
  1999.

\bibitem{balcan2007margin}
M.-F. Balcan, A.~Broder, and T.~Zhang, ``Margin based active learning,'' in
  \emph{International Conference on Computational Learning Theory}.\hskip 1em
  plus 0.5em minus 0.4em\relax Springer, 2007, pp. 35--50.

\bibitem{fails2003interactive}
J.~A. Fails and D.~R. Olsen~Jr, ``Interactive machine learning,'' in
  \emph{Proceedings of the 8th International Conference on Intelligent User
  Interfaces}.\hskip 1em plus 0.5em minus 0.4em\relax ACM, 2003, pp. 39--45.

\bibitem{verma2012learning}
N.~Verma, D.~Mahajan, S.~Sellamanickam, and V.~Nair, ``Learning hierarchical
  similarity metrics,'' in \emph{Proceedings of the IEEE Conference on Computer
  Vision and Pattern Recognition, CVPR}.\hskip 1em plus 0.5em minus 0.4em\relax
  IEEE, 2012, pp. 2280--2287.

\bibitem{ordonez2013large}
V.~Ordonez, J.~Deng, Y.~Choi, A.~C. Berg, and T.~L. Berg, ``From large scale
  image categorization to entry-level categories,'' in \emph{Proceedings of the
  IEEE International Conference on Computer Vision, ICCV}.\hskip 1em plus 0.5em
  minus 0.4em\relax IEEE, 2013, pp. 2768--2775.

\bibitem{Kuang2018Integrating}
Z.~Kuang, J.~Yu, Z.~Li, B.~Zhang, and J.~Fan, ``Integrating multi-level deep
  learning and concept ontology for large-scale visual recognition,''
  \emph{Pattern Recognition}, vol.~78, pp. 198--214, 2018.

\bibitem{Griffin2008Learning}
G.~Griffin and P.~Perona, ``Learning and using taxonomies for fast visual
  categorization,'' in \emph{Proceedings of the IEEE Conference on Computer
  Vision and Pattern Recognition, CVPR}, 2008, pp. 1--8.

\bibitem{Liu2013Probabilistic}
B.~Liu, F.~Sadeghi, M.~Tappen, O.~Shamir, and C.~Liu, ``Probabilistic label
  trees for efficient large scale image classification,'' in \emph{Proceedings
  of the IEEE Conference on Computer Vision and Pattern Recognition, CVPR},
  2013, pp. 843--850.

\bibitem{Fan2015Hierarchical}
J.~Fan, J.~Peng, L.~Gao, and N.~Zhou, ``Hierarchical learning of tree
  classifiers for large-scale plant species identification,'' \emph{IEEE
  Transactions on Image Processing, TIP}, vol.~24, no.~11, pp. 4172--4184,
  2015.

\bibitem{Dong2013Training}
P.~Dong, K.~Mei, N.~Zheng, H.~Lei, and J.~Fan, ``Training inter-related
  classifiers for automatic image classification and annotation,''
  \emph{Pattern Recognition}, vol.~46, no.~5, pp. 1382--1395, 2013.

\bibitem{Qu2017Joint}
Y.~Qu, L.~Lin, F.~Shen, C.~Lu, Y.~Wu, Y.~Xie, and D.~Tao, ``Joint hierarchical
  category structure learning and large-scale image classification,''
  \emph{IEEE Transactions on Image Processing, TIP}, vol.~26, no.~9, pp.
  4331--4346, 2017.

\bibitem{Zheng2017Hierarchical}
Y.~Zheng, J.~Fan, J.~Zhang, and X.~Gao, ``Hierarchical learning of multi-task
  sparse metrics for large-scale image classification,'' \emph{Pattern
  Recognition}, vol.~67, no.~C, pp. 97--109, 2017.

\bibitem{wang2019learning}
H.~Wang, S.~Ge, E.~P. Xing, and Z.~C. Lipton, ``Learning robust global
  representations by penalizing local predictive power,'' \emph{arXiv preprint
  arXiv:1905.13549}, 2019.

\bibitem{ren2018meta}
M.~Ren, E.~Triantafillou, S.~Ravi, J.~Snell, K.~Swersky, J.~B. Tenenbaum,
  H.~Larochelle, and R.~S. Zemel, ``Meta-learning for semi-supervised few-shot
  classification,'' \emph{arXiv preprint arXiv:1803.00676}, 2018.

\end{thebibliography}

\end{document}